# Probabilistic Disjunctive Logic Programming*

Liem Ngo
Decision Systems and Artificial Intelligence Lab
Department of Electrical Engineering
and Computer Science
University of Wisconsin-Milwaukee
Email: liem@cs.uwm.edu

## Abstract

In this paper we propose a framework for combining Disjunctive Logic Programming and Poole's Probabilistic Horn Abduction. We use the concept of hypothesis to specify the probability structure. We consider the case in which probabilistic information is not available. Instead of using probability intervals, we allow for the specification of the probabilities of disjunctions. Because minimal models are used as characteristic models in disjunctive logic programming, we apply the principle of indifference on the set of minimal models to derive default probability values. We define the concepts of *explanation* and *partial explanation* of a formula, and use them to determine the default probability distribution(s) induced by a program. An algorithm for calculating the default probability of a goal is presented.

## 1 INTRODUCTION

Two main approaches to coping with uncertainty in logic programming and deductive databases are disjunctive logic programming and quantitative logic programming. While disjunctive logic programming [Lobo et al., 1992] expresses uncertainty by using the indefiniteness inherent in disjunction, quantitative logic programming represents uncertainty by associating numerical quantities with clauses [Ng and Subrahmanian, 1992]. To our knowledge, there has been no effort to combine the disjunctive and quantitative approaches in a single logic programming framework.

A disjunctive logic program (or disjunctive deductive database) is characterized by its set of minimal models [Minker, 1982], where each model is conceived of as a possible state of the world. Traditional disjunctive logic programming semantics does not assign a preference or likelihood ranking to the states. But the ability to express preferences among possible states is crucial for many kinds of reasoning, such as abductive reasoning in diagnosis problems. Given a disjunctive logic program, we propose quantifying beliefs in facts by assigning more weight to facts that are true in a larger number of minimal models of the program.

Among the frameworks for quantitative logic programming, those based on probability theory have the most solid semantic foundation and the greatest potential for application. But the probabilistic approach suffers from the data collection problem. Usually, complete probability information is hard to obtain and experts often disagree on the exact probability values. For these reasons, it can be desirable to have a method of reasoning that does not require as input a complete specification of a probability distribution. One common approach is to reason with probability intervals [Ramoni, 1995]. A second approach is to use the principle of maximum entropy [Jaynes, 1979] to complete a partially specified distribution. In this paper, we investigate the latter approach in the context of disjunctive logic programming.

We propose a probabilistic disjunctive logic programming framework that allows for the expression of both probabilistic uncertainty and indefiniteness in the same program. The framework, which is an extension of Poole's Probabilistic Horn Abduction [Poole, 1993] and of disjunctive logic programming [Lobo et al., 1992], provides a natural representation of partial probabilistic information. In this initial attempt we confine ourselves to positive disjunctive logic programs [Lobo et al., 1992]. We give a semantics to the new language which extends the *minimal model* semantics [Lobo et al., 1992] for disjunctive logic programming and *possible world* semantics [Nilsson, 1986] for probability logics. A program in our framework is characterized by a probability distribution on *possible subspaces*. Each subspace is a set of minimal models and we use the *principle of indifference* [Jaynes, 1979] to assign probabilities to each minimal model in the subspace (if the number of minimal models is finite). We present a procedure to compute the probabilities of ground formulas and investigate its properties.

*This work is supported in part by a UWM graduate school fellowship and by NSF grant IRI-9509165.

In Section 2 we review the concepts of disjunctive logic programming and its minimal model semantics. Section 3 introduces the concepts of hypothesis and probabilistic disjunctive logic programming. We address the probabilistic semantics in the following two sections. We consider the case in which the hypothesis universe is finite in Section 4. In the general case, we use the concepts of full explanations and partial explanations to characterize the class of finitely defined formulas and show how to find the default probability of such formulas. We introduce the concepts of hypothetical model trees and forests as the main representation structures of a query-answering procedure in Section 6. Because of space limitation, proofs are omitted.

## 2 DISJUNCTIVE LOGIC PROGRAMS

As a common convention in logic programming, all models we mention in this paper are Herbrand models.

**Definition 1** *A* **disjunctive logic program clause** *is a formula of the form:* $A_1 \vee \ldots \vee A_k \leftarrow B_1, \ldots, B_m$, *where* $k \geq 1, m \geq 0$ *and* $A_1, \ldots, A_k, B_1, \ldots, B_m$ *are atoms. The part on the left of* $\leftarrow$ *is called the* **head** *and the other part is called the* **body** *of the clause. A* **disjunctive logic program (DLP)** *is a finite set of disjunctive logic program clauses. A DLP without function symbols is called a* **disjunctive deductive database (DDB)**.

**Example 1** *Consider the following hypothetical DLP:*
$P = \{fac(X) \vee staff(X) \leftarrow work(X, uwm);$
$doc(X) \vee fac(X) \vee staff(X) \leftarrow work(X, mcw); work(X, uwm) \leftarrow dad(X, bob);$
$work(X, mcw) \leftarrow dad(X, helen); dad(alex, helen) \vee dad(alex, bob)\}$

*In this example we assume that alex is either the father of helen or bob. The father of helen works for mcw— the Medical College of Wisconsin. Bob's father works for uwm—the University of Wisconsin-Milwaukee. An employee of mcw may be a staff or faculty or doctor, while an employee of uwm may only be a staff or faculty.*

The meaning of a DLP is usually characterized by its set of minimal models. A ground formula $F$ is considered a logical consequence of a DLP $P$ if $F$ is evaluated to *true*, in the usual sense, in all minimal models of $P$. [1]

**Example 2** *The program in the above example is characterized by the following set of minimal models:*

---
[1] Minimal model semantics interprets disjunctions as representing exclusive or. Although this interpretation may be too restrictive, we use it in this paper since it is the most popular approach to interpreting disjunctive logic programs.

$\{fac(alex), work(alex, uwm), dad(alex, bob)\};$
$\{staff(alex), work(alex, uwm), dad(alex, bob)\};$
$\{doc(alex), work(alex, mcw), dad(alex, helen)\};$
$\{fac(alex), work(alex, mcw), dad(alex, helen)\};$
$\{staff(alex), work(alex, mcw), dad(alex, helen)\}$

*Taking this set of minimal models as the meaning of the program, it has the followings as some of its logical consequences:* $work(alex, uwm) \vee work(alex, mcw)$, $fac(alex) \vee doc(alex) \vee staff(alex)$ *and* $\neg dad(alex, bob) \vee \neg dad(alex, helen)$. *The last disjunction demonstrates the exclusivity property of minimal model semantics: Alex cannot be both the father of Helen and Bob.*

Minimal models are usually considered as representing possible states of the world [Minker, 1982]. If we would like to assign degrees of belief to ground formulas, the set of minimal models forms a reasonable sample space. But we need a method of assigning probabilities to the elements of the sample space. A common approach is to use the principle of maximum entropy: the belief in a ground formula $F$ is equal to the ratio of the number of minimal models in which $F$ is true over the total number of minimal models. In the above example, we should assign a probability of $1/5$ to $doc(alex)$. In related work, Grove et al. [1994] apply the principle of maximum entropy to the set of all possible models for a set of sentences in probability logic. We hope that by using semantic models the problem of language dependence [Grove et al., 1994] associated with the principle of maximum entropy can be alleviated. There are proposals, e.g. [Sakama and Inoue, 1993], to use some nonminimal models as characteristic models of a DLP. Our semantic framework and procedure can be easily adapted to these extensions.

## 3 PROBABILISTIC DISJUNCTIVE LOGIC PROGRAMMING

Poole [1993] introduces an abductive framework for incorporating probabilistic reasoning into Horn logic programs. Probabilistic information is encoded in a probability distribution on a specific set of ground atoms called hypotheses. Hypotheses are divided into disjoint finite sets, each set is declared by a ground instance of a *disjoint declaration*.

**Definition 2** *([Poole, 1993])* A **disjoint declaration** *is a declaration of the form:* $disjoint(h_1 : p_1, \ldots, h_n : p_n)$ *where* $h_i$ *are atoms,* $n > 1$, $p_i \geq 0$ *and* $p_1 + \cdots + p_n = 1$. *The meaning is if* $h'$ *is a ground instance of* $h_i$ *then the probability that* $h'$ *is true is* $p_i$. *Each* $h_i$ *is called a* **hypothesis** *or* **assumable**. *We call a ground instance of a hypothesis a* **ground hypothesis**.

The **hypothesis universe**, denoted by $\mathcal{H}_P$, is the set of all ground instances of the hypotheses that appear in the disjoint declarations. We call the non-hypothesis atoms **regular atoms**.

Each ground instance of a disjoint declaration represents the possible realizations of a random variable—each hypothesis in the sentence represents a possible state of the random variable and each number characterizes the chance of the corresponding state. The reasoning in such a framework is simplified by the assumption that all random variables are mutually independent.

**Definition 3** *A* **probabilistic disjunctive logic program (PDLP)** *P consists of the following components: (1) A set $DS_P$ of disjoint statements such that no two different ground instances of disjoint statements share common hypothesis(es). (2) A DLP $DP_P$ such that no ground instance of a hypothesis appears in the head of a ground instance of a clause in $DP_P$.*

We take such a declaration of $P$ as shorthand for saying that $P$ has the following components:

(1) The DLP $DP_P$.

(2) A set of integrity constraints $IC_P$ which is formed in the following way: If the disjoint declaration is $disjoint(h_1 : p_1, \ldots, h_n : p_n)$ then form the set $\{h_1 \vee \ldots \vee h_n\} \cup \{\leftarrow h_i, h_j | 1 \leq i \neq j \leq n\}$, where all variables are assumed to be universally quantified over the entire clause. $IC_P$ is the union of all such sets.

(3) A probability distribution $Pr^*$ specified by: (a) If $disjoint(h'_1 : p_1, \ldots, h'_n : p_n)$ is a ground instance of a disjoint declaration then the probability that $h'_i$ is *true*, denoted by $Pr^*(h'_i)$, is $p_i$. (b) The probabilistic independence assumption: if $H = \{h_1, \ldots, h_m\}, m > 1$, is a set of ground hypotheses such that $H \cup IC_P$ is logically consistent then $Pr^*(h_1 \wedge \cdots \wedge h_m) = Pr^*(h_1) \times \cdots \times Pr^*(h_m)$.

In the remainder of the paper we say a set of ground hypotheses $H$ is **consistent** if the set $H \cup IC_P$ is logically consistent.

The function $Pr^*$ can be extended to be a probability assignment to all finite propositional formulas $F$ containing only ground hypotheses in the following way:

(P1) Convert $F$ into a formula of the form $\bigvee_i \bigwedge_j h_{ij}$, where the $h_{ij}$ are ground hypotheses by replacing $\neg h_i$ by $\bigvee_{j \neq i} h_j$ if $disjoint(h_1 : \alpha_1, \ldots, h_n : \alpha_n)$ is a ground instance of a disjoint statement containing $h_i$.

(P2) Evaluate the resulting formula by using the following rules: (1) $Pr^*(h_1 \wedge \cdots \wedge h_m) = Pr^*(h_1) \times \cdots \times Pr^*(h_m)$, if $\{h_1, \ldots, h_m\}$ is consistent. (2) $Pr^*(h_1 \wedge \cdots \wedge h_m) = 0$, if $\{h_1, \ldots, h_m\}$ is not consistent. (3) $Pr^*(F_1 \vee F_2) = Pr^*(F_1) + Pr^*(F_2) - Pr^*(F_1 \wedge F_2)$.

In the succeeding sections we extend $Pr^*$ to formulas involving regular atoms.

**Example 3** *To illustrate various techniques and definitions in this paper, we extend Example 1 by introducing probabilities. Suppose the following statistics are available: (1) We know that a person working for uwm may be a staff or a faculty, but do not know the ratio of staff and faculty. (2) We know that 20% of mcw employees have doctoral degrees and those people work either as a doctor or a faculty. (3) The mcw employees who do not have doctoral degrees work as staff. (4) There is a .7 chance that Alex is either the father of Bob or Helen.*

*The PDLP P incorporating the above partial probabilistic information is:*
$DP_P = \{$
$(1) fac(X) \vee staff(X) \leftarrow work(X, uwm);$
$(2) doc(X) \vee fac(X) \leftarrow work(X, mcw), hasDoc;$
$(3) staff(X) \leftarrow work(X, mcw), noDoc;$
$(4) work(X, uwm) \leftarrow dad(X, bob);$
$(5) work(X, mcw) \leftarrow dad(X, helen);$
$(6) dad(alex, helen) \vee dad(alex, bob) \leftarrow haveRel\}.$
$DS_P = \{disjoint(hasDoc : .2, noDoc : .8);$
$disjoint(haveRel : .7, noRel : .3)\}.$

*The hypotheses are hasDoc, noDoc, haveRel and noRel. The hypothesis universe $\mathcal{H}_P$ is $\{hasDoc, noDoc, haveRel, noRel\}$.*

$Pr^*(hasDoc \wedge haveRel) = .2 \times .7 = .14; Pr^*(hasDoc \vee haveRel) = Pr^*(hasDoc) + Pr^*(haveRel) - Pr^*(hasDoc \wedge haveRel) = .2 + .7 - .14 = .76 = Pr^*(hasDoc \wedge haveRel) + Pr^*(hasDoc \wedge noRel) + Pr^*(noDoc \wedge haveRel) = 1 - Pr^*(noDoc \wedge noRel).$

Let $H = \{h_1, \ldots, h_n\}$ be a consistent set of ground hypotheses. The completion of $H$, $compl(H)$, is the set of all $H'$ that can be formed from $H$ by substituting for each $h_i$ a ground hypothesis from its disjoint statement. If $HS$ is a set of consistent sets of hypotheses, the **completion** of $HS$, denoted by $COMPL(HS)$, is defined as $compl(H^*)$, where $H^*$ is a maximal consistent subset of $\cup_{H \in HS} H$. We also define the **expansion** of $HS$, denoted by $expd(HS)$, as the set of all elements in $COMPL(HS)$ which contain an element of $HS$.

**Example 4** *Continuing Example 3, we assume that $H = \{hasDoc, haveRel\}$ and $HS = \{\{noDoc\}, \{haveRel\}\}$. Then, $compl(H)$ is the set of all hypothesis bases, $COMPL(HS) = compl(H)$ and $expd(HS) = \{\{hasDoc, haveRel\}, \{noDoc, haveRel\}, \{noDoc, noRel\}\}.$*

**Definition 4** *A* **possible model** *of a PDLP P is a* **minimal model** *of the corresponding set of sentences $DP_P \cup IC_P$. We denote by $PW_P$ the set of all possible models of P.*

A possible model plays the same role as a possible world in probabilistic logic [Nilsson, 1986]. Hence, we usually use $w$ to designate a possible model and $W$ to designate a set of possible models. Let $F$ be a ground formula, we denote by $W(F)$ the set of all possible models of $P$ such that $F$ is true, in the usual sense, in each of them.

Because $IC_P$ states that the hypotheses in each disjoint statement are mutually exclusive and exhaustive, each possible model $w$ contains one and only one hypothesis from each ground instance of a disjoint declaration. A set of hypotheses formed in this way is called a hypothesis base.

**Definition 5** *Let $P$ be a PDLP. A* **hypothesis base** *is a maximal consistent set of hypotheses. The* **basic subspace** *corresponding to a hypothesis base $H$, denoted by $B_H$, is the set of all minimal models of $DP_P \cup H \cup IC_P$.*

**Example 5** *In the example 3, $H = \{hasDoc, haveRel\}$ is a hypothesis base. Let*
$M_1 = H \cup \{fac(alex), work(alex, uwm), dad(alex, bob)\}$;
$M_2 = H \cup \{staff(alex), work(alex, uwm), dad(alex, bob)\}$;
$M_3 = H \cup \{doc(alex), work(alex, mcw), dad(alex, helen)\}$;
$M_4 = H \cup \{fac(alex), work(alex, mcw), dad(alex, helen)\}$.
*Then $B_H = \{M_1, M_2, M_3, M_4\}$.*

It is easy to see from Definition 5 that the set of hypotheses contained in each $w \in B_H$ is exactly $H$. Hence, the set $\{B_H | H$ is a hypothesis base$\}$ forms a partition of $PW_P$: (1) if $H$ and $H'$ are two different hypothesis bases then $B_H \cap B_{H'} = \{\}$; (2) $PW_P = \bigcup_H B_H$.

## 4 PROBABILISTIC SEMANTICS: THE CASE OF FINITE HYPOTHESIS UNIVERSE

In this section we consider the case where the hypothesis universe $\mathcal{H}_P$ of a program $P$ is finite. Notice that the language may contain function symbols, in which case the Herbrand base is infinite. Because $\mathcal{H}_P$ is finite, each hypothesis base is finite. Since $B_H$ is the set of minimal models that include the hypothesis base $H$, we can assign to each basic subspace $B_H$ the probability of $H$:

$(Pr1) \qquad Pr(B_H) = \prod_{h \in H} Pr^*(h)$

**Example 6** *Continuing Example 5, $H = \{hasDoc, haveRel\}$ is a hypothesis base. $Pr(B_H) = Pr^*(hasDoc) \times Pr^*(haveRel) = .2 \times .7 = .14$. We can see that $M_1, M_2, M_3$ and $M_4$ are not assigned a probability by rule $(Pr1)$.*

Because $M_1, M_2, M_3$ and $M_4$ are the only possible models under the assumption $haveRel$ and $hasDoc$, we can use the principle of indifference [Jaynes, 1979] to assign equal probability to $M_1, M_2, M_3$ and $M_4$ (i.e. $.14/4 = .035$). In general, if $B$ is a basic subspace then every proper subset of $B$ is not assigned a probability by $(Pr1)$. If $B$ is a finite basic subspace, we should assign the probability $Pr(B)/card(B)$ to each possible model $w \in B$, where $card(B)$ is the number of elements in $B$.

We are concerned with computing $Pr(F)$, where $F$ is an arbitrary ground formula. The set of possible models in which $F$ is true, $W(F)$, can be divided into two parts:

(F1) The first part is a union of some basic subspaces $\bigcup_{H \in I} B_H$, where $I$ is a set of hypothesis bases. This part can be assigned the probability

$$\sum_{H \in I} Pr(B_H) = \sum_{H \in I} \prod_{h \in H} Pr^*(h)$$

(F2) The second part is a union of portions of basic subspaces. If each possible model in this part can be assigned a probability by the principle of indifference then their sum can be used as the probability of the second part.

The sum of the above two values can be used as the *default probability* of $F$. Each ground formula which can be assigned a default probability is called *measurable*.

## 5 PROBABILISTIC SEMANTICS: THE GENERAL CASE

In this section, we consider the case in which the language contains function symbols and the hypothesis universe $\mathcal{H}_P$ may be infinite. If $\mathcal{H}_P$ is infinite then each hypothesis base is infinite and we cannot use rule $(Pr1)$ to compute the probability of a basic subspace. We need to bundle the basic subspaces into measurable sets (corresponding to finite sets of hypotheses).

In order to compute $Pr(F)$, where $F$ is a ground formula, we divide $W(F)$ into two parts. The first part is "fully explainable" by hypotheses (corresponding to rule (F1) above). The second part is "partially explainable" by hypotheses (corresponding to rule (F2) above) and the principle of indifference is used to derive a default probability measure. We generalize the concept of explanation in [Poole, 1993] to that of full explanation and partial explanation.

It is easy to see that if $H$ and $H'$ are two sets of hypotheses and $H \subseteq H'$ then every minimal model of $DP_P \cup H' \cup IC_P$ is a minimal model of $DP_P \cup H \cup IC_P$. In particular, if $H$ is a finite consistent set of hypotheses then the set of minimal models of $DP_P \cup H \cup IC_P$ is exactly $\cup \{B_{H'} | H'$ is a hypothesis base superset of $H\}$.

### 5.1 EXPLANATIONS

Let $F$ be a ground formula. A **full explanation (f-explanation)** of $F$ (wrt the PDLP $P$) is a consistent set of ground hypotheses $H$ such that $F$ is true in every minimal model of $DP_P \cup H \cup IC_P$.

By the above observation, if $H$ is an f-explanation of $F$ then a consistent hypothesis superset $H'$ of $H$, is also an f-explanation of $F$. Because we define the probability of a formula by a finite number of arithmetic operations, we want to consider only finite f-explanations. In fact, we concentrate on finite minimal f-explanations of $F$.

**Definition 6** *Let $F$ be a ground formula. A **minimal f-explanation** of $F$ (wrt the PDLP $P$) is an f-explanation $H$ of $F$ such that every proper subset of $H$ is not an f-explanation of $F$. We call the set of all minimal f-explanations of $F$ its **f-explanation base** and denote it by $f\text{-}expl(F)$.*

*We say $F$ is **finitely f-explainable** if (1) it has a finite number of minimal f-explanations, and (2) each minimal f-explanation of $F$ is finite.*

Let $F$ be a finitely f-explainable ground formula. We want to evaluate the probability of $W$, the set of all possible models resulting from $f\text{-}expl(F)$. Let $f\text{-}expl(F) = \{\{h_{i1}, \ldots, h_{in_i}\} | i = 1, \ldots, m\}$. From $f\text{-}expl(F)$ we form the formula $F^* = \bigvee_{i=1}^{m} \bigwedge_{j=1}^{n_i} h_{ij}$. We define $Pr_{full}(F)$ as $Pr^*(F^*)$, which is defined in Section 3.

### 5.2 PARTIAL EXPLANATIONS

For a ground formula $F$, besides the f-explanations, there are sets of hypotheses such that $F$ is true in only some minimal models resulting from them. We call these sets **partial explanations**.

Let $F$ be a ground formula. A **partial explanation (p-explanation)** of $F$ is a consistent set of ground hypotheses $H$ such that:

(1) $H$ is not a subset of any f-explanation of $F$.

(2) $F$ is true in at least one minimal model and false in at least one minimal model of $DP_P \cup H \cup IC_P$.

Condition 1 eliminates artificial partial explanations which are formed by simply dropping hypotheses from f-explanations.

**Definition 7** *A p-explanation $H$ of $F$ is **sufficient** if for every consistent set of ground hypotheses $H'$ such that $H \subseteq H'$, $\{w - \mathcal{H}_P | w \text{ is a minimal model of } DP_P \cup H' \cup IC_P\} = \{w - \mathcal{H}_P | w \text{ is a minimal model of } DP_P \cup H \cup IC_P\}$.*

*A p-explanation $H$ of $F$ is **minimally sufficient** if there is no proper subset $H'$ of $H$ which is also a sufficient p-explanation of $F$.*

*The **p-explanation base** of $F$, denoted by $p\text{-}expl(F)$, is the set of all minimally sufficient p-explanations of $F$.*

A p-explanation is sufficient if any superset of it generates essentially the same set of minimal models.

It is obvious from the above definition that: (1) For any $F$, its p-explanation base exists and is unique. (2) If $H$ is a sufficient p-explanation then any consistent hypothesis superset of $H$ is also a sufficient p-explanation.

**Definition 8** *A p-explanation base is called **finite** if (1) it is a finite set; (2) it contains only finite p-explanations; and (3) if $H$ is one of its element then the set $\{w - \mathcal{H}_P | w \text{ is a minimal model of } DP_P \cup H \cup IC_P\}$ is finite.*

Condition (3) is necessary for the application of the principle of indifference. We need to eliminate the hypotheses from each $w$ because there may be an infinite number of hypothesis base supersets of $H$. The conditions (1) and (2) allow us to compute the probability by finite sums and finite products.

**Example 7** *(1) Consider the program $P$ in which $DP_P = \{a \vee b \leftarrow h_1; a \vee c \leftarrow h_2\}$, $DS_P = \{disjoint(h_1 : .5, h_1' : .5); disjoint(h_2 : .5, h_2' : .5)\}$.*

*The two sets $\{h_1\}$ and $\{h_2\}$ are not sufficient partial explanations of $a$. The finite p-explanation base of $a$ is $ES = \{\{h_1, h_2\}, \{h_1, h_2'\}, \{h_1', h_2\}\}$. Every member in that set is a minimally sufficient p-explanation. In this case, $expd(ES) = ES$.*

*(2) Consider the program $P'$ in which $DP_{P'} = \{d \vee e \leftarrow h_3; d \vee e \leftarrow h_4\}$ and $DS_{P'} = \{disjoint(h_3 : .5, h_3' : .5); disjoint(h_4 : .4, h_4' : .6)\}$.*

*The two sets $\{h_3\}$ and $\{h_4\}$ are minimally sufficient partial explanations of $d$. The finite p-explanation base of $d$ is $ES = \{\{h_3\}, \{h_4\}\}$. The expansion of $ES$ is $expd(ES) = \{\{h_3, h_4'\}, \{h_3, h_4\}, \{h_3', h_4\}\}$.*

Let $F$ be a ground formula. We say $F$ is **finitely p-explainable** if its p-explanation base is finite.

**Proposition 1** *Let $F$ be a ground formula. (1) If the language does not contain function symbols then $F$ is finitely p-explainable. (2) Each p-explanation of $F$ is inconsistent with each f-explanation of $F$.*

Let $F$ be a finitely p-explainable ground formula. We want to evaluate the probability of the set of all possible models resulting from $p\text{-}expl(F)$ and satisfying $F$. In the first step, we form the expansion of $p\text{-}expl(F)$. Assume $expd(p\text{-}expl(F)) = \{\{h_{i1}, \ldots, h_{in}\} | i = 1, \ldots, m\}$.

Let $H$ be an element of $expd(p\text{-}expl(F))$. The probability of $H$ can be computed as $\prod_{h \in H} Pr^*(h)$. Let $W$ be the set $\{w - \mathcal{H}_P | w \text{ is a minimal model of } DP_P \cup H \cup IC_P\}$. Because $W$ is finite we can count the number of elements. Let $m_H$ denote the number of elements and let $m_H^F$ denote the number of elements that satisfy $F$. A reasonable probability weight for the p-explainable portion of $W(F)$ is $\sum_{H \in expd(pexpl(F))}(\prod_{h \in H} Pr^*(h) \times m_H^F/m_H)$, which we denote by $Pr_{partial}(F)$.

**Example 8**
*In Example 7.(2), the finite p-explanation base of d is p-expl(F) = $\{\{h_3\}, \{h_4\}\}$. The expansion of p-expl(F) is expd(p-expl(F)) = $\{H_1, H_2, H_3\}$, where $H_1 = \{h_3, h_4'\}, H_2 = \{h_3, h_4\}$, and $H_3 = \{h_3', h_4\}$.*

*There are two sets in $\{w - \mathcal{H}_P | w$ is a minimal model of $DP_P \cup H_1 \cup IC_P\}$ : $\{\epsilon\}$ and $\{d\}$. Similarly, half of the sets resulting from $H_2$ (and $H_3$) contain d.*

$Pr_{partial}(d) = \sum_{i=1}^{3}(\prod_{h \in H_i} Pr^*(h) \times .5) = .5 \times .4 \times .5 + .5 \times .6 \times .5 + .5 \times .4 \times .5 = .35.$

### 5.3 THE DEFAULT PROBABILITY DISTRIBUTION

Let $P$ be a PDLP. A ground formula $F$ is called **finitely defined** if it is both finitely f-explainable and finitely p-explainable.

**Proposition 2** *If $F_1$ and $F_2$ are finitely defined ground formulas then the following formulas are finitely defined: $\neg F_1$, $F_1 \wedge F_2$, and $F_1 \vee F_2$.*

If $F$ is finitely defined then the *default probability* of $F$ is defined as

$$Pr(F) = Pr_{full}(F) + Pr_{partial}(F),$$

where $Pr_{full}(F)$ and $Pr_{partial}(F)$ are defined in the previous sections.

The following proposition shows that $Pr()$ satisfies the properties of a probability distribution.

**Proposition 3** *If $F_1$ and $F_2$ are finitely defined ground formulas then (1) $Pr(\neg F_1) = 1 - Pr(F_1)$, (2) $Pr(F_1 \wedge F_2) + Pr(F_1 \vee F_2) = Pr(F_1) + Pr(F_2)$, and (3) $Pr(True) = 1, Pr(False) = 0$.*

## 6 A BOTTOM-UP PROCEDURE

In this section we propose a procedure to compute $Pr(F)$ by generating the minimal models under various hypothesis sets. We generalize the concept of **model trees** [Lobo *et al.*, 1992], which is used to represent the set of minimal models of a disjunctive deductive database, and use the new concept to represent the possible models of a **probabilistic disjunctive deductive database (PDDB)**. Because a PDDB does not contain function symbols, there are a finite number of minimal models and each minimal model is finite.

The following definition is adapted from Fernandez and Minker [1991].

**Definition 9** *Let $\Im$ be a finite set of Herbrand interpretations (models) of a DLP $P$. An interpretation (model) tree for $\Im$ is a tree structure where (1) The root is labeled by the special symbol $\epsilon$. Other nodes are labeled with atoms in $\Im$ or $\epsilon$. (2) No atom occurs more than once in a path from the root to the leaf node. Each such a path is called a branch. (3) $I \in \Im$ iff there exists a branch b of the tree such that $I = \{A | A \neq \epsilon$ and $A$ appears in $b\}$.*

```
INPUT: A hypothetical model forest ⟨⟨MT₁, H₁⟩, ...,
⟨MTₙ, Hₙ⟩⟩ and a ground formula F.
OUTPUT: Pr(F).
Let       Prob := 0;
For       i := 1 to n do
          begin Let p := the number of minimal
                    models in MTᵢ satisfying F;
                Let q := the total number of minimal
                    models in MTᵢ;
                If q ≠ 0 then
                    Prob := Prob + (p/q) × Pr*(Hᵢ)
          end;
Output(Prob).
```

Figure 1: Computing $Pr(F)$ from a complete hypothetical model forest.

*If $\Im$ is the set of all minimal models of $P$, we call such a tree a minimal model tree of $P$. An empty model tree is a model tree containing only $\epsilon$ node.*

**Example 9** *If $P$ is the DDB $\{dad(alex, helen) \vee dad(alex, bob)\}$ then the model tree of $P$ is the tree $MT_1$ in Figure 3.(a).*

In our framework, the possible models are generated with respect to certain set of hypotheses. We represent the semantics structure of a PDDB by a set of pairs ⟨a set of hypotheses $H$, the possible models 'generated' from $H$⟩.

**Definition 10** *Let $P$ be a PDDB. A **hypothetical model forest** of $P$ is a list of $n, n \geq 0$, pairs $\langle MT_i, H_i \rangle$ such that:*

*(1) $H_i, i = 1, \ldots, n$, are consistent sets of ground hypotheses and each $MT_i$ is a minimal model (with hypotheses eliminated) tree of $DP_P \cup H_i \cup IC_P$.*

*(2) If $1 \leq i \neq j \leq n$ then $H_i \cup H_j$ is inconsistent.*

*A hypothetical model forest of $P$ is complete if for an arbitrary hypothesis base $H$ there exists a pair $\langle MT_i, H_i \rangle$ such that: (1) $H_i \subseteq H$; and (2) $\{w - H | w \in B_H\}$ is exactly the set of minimal models of $MT_i$.*

A complete hypothetical model forest contains the possible models under any hypothesis base.

**Proposition 4** *If $\langle \langle MT_1, H_1 \rangle, \ldots, \langle MT_n, H_n \rangle \rangle$ is a complete hypothetical model forest of a PDDB $P$ and $F$ is a ground formula of regular atoms then $Pr(F)$ can be computed using the algorithm in Figure 1.*

```
INPUT: A PDDB P.
OUTPUT: A complete hypothetical model forest of P.
Assume    P = {C_1,...,C_n};
Let       HMF := ⟨□,{}⟩;
Repeat
          For i := 1 to n do
              update(C_i, HMF);
Until no modification is performed on HMF;
Output(HMF).
```

Figure 2: Complete Hypothetical Minimal Model Forest construction procedure.

### 6.1 BUILDING A COMPLETE HYPOTHETICAL MODEL FOREST

In this section we present an adaptation of the minimal model tree building procedure presented by Fernandez and Minker [1991] to build a complete hypothetical model forest from a PDDB $P$.

The procedure incrementally constructs a complete hypothetical model forest by considering one database clause at a time. The main procedure in Figure 2 is an iterative process that proceeds until no new atoms can be added to the forest. In the procedure we use □ to indicate the empty model tree. Every clause in the given PDDB $P$ is used in turn to update the partially built forest.

**Example 10** *We want to show the construction of a complete hypothetical model forest for the PDDB in Example 3.*

*Assume that clause (6) is selected first. After processing (6) the variable HMF contains $\langle\langle MT_1, \{haveRel\}\rangle; \langle\square, \{noRel\}\rangle\rangle$ where $MT_1$ is the tree in Figure 3.a and □ is the empty tree. Next, assume that clauses (4) and (5) are selected, the resulting HMF is $\langle\langle MT_2, \{haveRel\}\rangle; \langle\square, \{noRel\}\rangle\rangle$, where $MT_2$ is shown in Figure 3.b. Now, clause (1) is selected and HMF becomes $\langle\langle MT_3, \{haveRel\}\rangle; \langle\square, \{noRel\}\rangle\rangle$. $MT_3$ is shown in Figure 3.c. If clause (2) is selected then the only non-empty element of HMF will be split into two pairs: HMF becomes $\langle\langle MT_3, \{haveRel, hasDoc\}\rangle; \langle MT_3, \{haveRel, noDoc\}\rangle; \langle\square, \{noRel\}\rangle\rangle$ before the first element is changed to $\langle MT_4, \{haveRel, hasDoc\}\rangle$. $MT_4$ is shown in Figure 3.d. Finally, clause (3) is selected and the minimal model tree of the second element of HMF is replaced by $MT_5$ (Figure 3.e).*

*The final HMF is $\langle\langle MT_4, \{haveRel, hasDoc\}\rangle; \langle MT_5, \{haveRel, noDoc\}\rangle; \langle\square, \{noRel\}\rangle\rangle$.*

Because of the space limitation, *update()* is not presented.

**Proposition 5** *The result returned by the procedure in Figure 2 is a complete hypothetical minimal model forest of $P$.*

The above procedure generates a complete hypothetical model forest from a PDDB. This strategy is appropriate if we want to preprocess the PDDB into a compact form which can be used to answer different queries. In the full paper we present a more efficient procedure to compute the default probability of a specific formula. That procedure has two phases. The first top-down phase determines the relevant hypothesis sets. The second phase uses that result to generate only a limited portion of a complete hypothetical model forest which is sufficient to compute the probability of the input formula.

## 7   APPLICATIONS AND RELATED WORK

**(1) Combining logic programs and probabilistic databases:** In a related paper, we show that the current framework can be easily extended to provide a probabilistic semantics to the coupling of disjunctive logic programs and Barbara et al.'s Probabilistic Databases [Barbara *et al.*, 1992].

**(2) Combining different ways of handling Null values in Databases:** Disjunctive and probabilistic formulations are two of the most important modes of Null value representation in database theory. The current work could offer a framework for combining the two approaches.

**(3) An extension of Poole's Probabilistic Abduction Theory:** Poole [1993] requires the theories in his framework to be acyclic and, hence, to have one and only one characteristic model. Only such theories are assigned a semantics. Our work removes that constraint. In Poole's framework, one simple way to handle probability of disjunction is applying the principle of indifference locally. That means each disjunctive rule $A_1 \vee \ldots \vee A_n \leftarrow body$ is represented by the set $\{A_i \leftarrow body, h_i; i = 1,\ldots,n\}$ and the statement $disjoint(h_1 : 1/n, \ldots, h_n : 1/n)$. Our proposal takes into account the interaction between rules and applies the principle of indifference globally.

**Example 11** *In one observation we know that Alex is either a laywer or a doctor (laywer $\vee$ doctor) and cannot be both. Another source confirms that Alex is either a laywer or professor (laywer $\vee$ professor) and cannot be both. Combining the two sources, we have the DLP $\{laywer \vee doctor; laywer \vee professor\}$. This DLP has two minimal models $\{laywer\}$ (Alex is a laywer) or $\{doctor, professor\}$ (Alex is both a doctor and a professor). Our proposal would assign .5 to the fact that Alex is a laywer. The solution in Poole's framework would introduce two disjoint statements $disjoint(ld : .5, ld' : .5)$ and $disjoint(lp : .5, lp' : .5)$ and represent the disjunctive information as the Horn program $\{lawyer \leftarrow ld; doctor \leftarrow ld'; lawyer \leftarrow lp; professor \leftarrow lp'\}$. In this formulation, the probability of Alex being a laywer is .75 and the probability that Alex being both a lawyer and a doctor is .25. The*

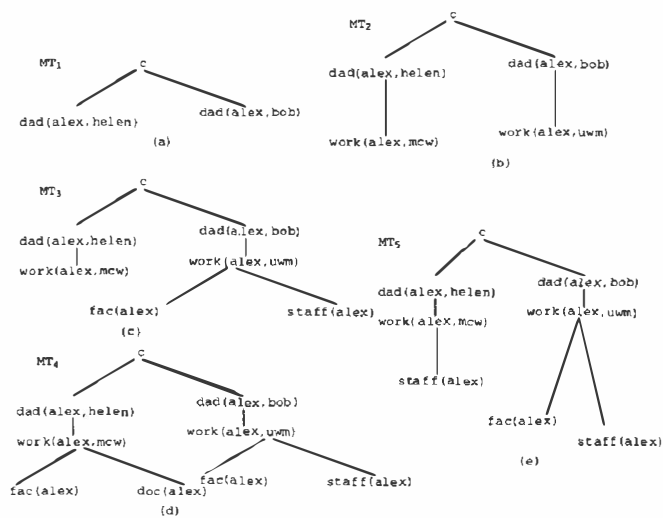

Figure 3: The non-empty trees in the generated complete hypothetical model forest.

*last fact contradicts the assumption that Alex cannot be both a lawyer and a doctor.*

## 8  FUTURE RESEARCH

We plan to extend the current probabilistic semantics to more advanced forms of disjunctive logic programming and to improve the inference procedures. In defining the default probability function, we have considered only point-valued probabilities. Hence, we restrict the p-explanations to generating a finite number of minimal models. We can extend further the default probability function by allowing for interval probabilities and accepting non-finite p-explanations as long as bounds on probabilities can be determined. The current procedure is inefficient in the sense that the entire minimal models must be generated. We conjecture that there might be some syntactic criteria which allow us to answer a given query using only some local portion of the given program.

### Acknowledgements

The author would like to thank his supervisor, Dr. Peter Haddawy, for his various help in the presentation of the paper and in other things. The comments of the anonymous reviewers are appreciated.